\documentclass[letterpaper]{article}
\pdfoutput=1
\usepackage{aaai}
\usepackage{times}
\usepackage{helvet}
\usepackage{courier}
\usepackage{graphicx}
\usepackage{listings}
\usepackage{wrapfig}
\usepackage{appendix}
\usepackage{caption}
\usepackage{subcaption}
\usepackage{lipsum}% http://ctan.org/pkg/lipsum
\usepackage{graphicx}% http://ctan.org/pkg/graphicx
\usepackage{comment}
\usepackage{xcolor}
\usepackage[normalem]{ulem}
\usepackage{hyperref}

\definecolor{codegreen}{rgb}{0,0.6,0}
\definecolor{codegray}{rgb}{0.3,0.3,0.3}
\definecolor{codepurple}{rgb}{0.58,0,0.82}
\definecolor{backcolour}{rgb}{0.98,0.98,0.97}
\definecolor{azure(colorwheel)}{rgb}{0.0, 0.5, 1.0}
\definecolor{goldenpoppy}{rgb}{0.99, 0.76, 0.0}
\definecolor{green(ryb)}{rgb}{0.4, 0.69, 0.2}

\lstdefinestyle{mystyle}{
    backgroundcolor=\color{white},   
    commentstyle=\color{codegreen},
    keywordstyle=\color{azure(colorwheel)},
    numberstyle=\tiny\color{codegray},
    stringstyle=\color{green(ryb)},
    basicstyle=\ttfamily\footnotesize,
    breakatwhitespace=false,         
    breaklines=true,                 
    captionpos=b,                    
    keepspaces=true,                 
    numbers=left,                    
    numbersep=5pt,                  
    showspaces=false,                
    showstringspaces=false,
    tabsize=2,
    showtabs=true
}
\definecolor{maroon}{rgb}{.5,0,0}
\definecolor{deepForestGreen}{rgb}{.2,.8,.04}
\definecolor{magenta}{rgb}{0.7,0,1}

\begin{document}
%\title{Fiber: Efficient Development and Distributed Training for Reinforcement Learning and Population-Based Methods}
\title{Fiber: A Platform for Efficient Development and Distributed Training for Reinforcement Learning and Population-Based Methods}

\author{Jiale Zhi\textsuperscript{1}\footnote{} \hspace{5mm} Rui Wang\textsuperscript{1} \hspace{5mm}  Jeff Clune\textsuperscript{2} \hspace{5mm} Kenneth O. Stanley\textsuperscript{1}\\
%\textsuperscript{1}{Uber AI} \hspace{5mm}
%\textsuperscript{2}{OpenAI} \\
%Uber AI\\
%San Francisco, CA 94103 \\
\texttt{{\{jiale,ruiwang\}}@uber.com} \hspace{2mm} \texttt{jeffclune@openai.com} \hspace{2mm} \texttt{kstanley@uber.com}\\
}

\maketitle
\begin{abstract}
\begin{quote}
  Recent advances in machine learning are consistently enabled by increasing amounts of computation.  Reinforcement learning (RL) and population-based methods in particular pose unique challenges for efficiency and flexibility to the underlying distributed computing frameworks.  These challenges include frequent interaction with simulations, the need for dynamic scaling, and the need for a user interface with low adoption cost and consistency across different backends. In this paper we address these challenges while still retaining development efficiency and flexibility for both research and practical applications by introducing Fiber, a scalable distributed computing framework for RL and population-based methods.  Fiber aims to significantly expand the accessibility of large-scale parallel computation to users of otherwise complicated RL and population-based approaches without the need to for specialized computational expertise.
\end{quote}
\end{abstract}

\section{Introduction}

Increasing computation underlies many recent advances in machine learning \cite{amodei_hernandez_2018}. More and more algorithms exploit parallelism and rely on distributed training for processing enormous amount of data, both in the conventional supervised context and in reinforcement learning (RL) \cite{espeholt2018impala,mnih2016asynchronous,ecoffet2019go,nair2015massively,horgan2018distributed,jaderberg2018human,sutton2018reinforcement}
and population-based methods \cite{salimans2017evolution,jaderberg2017population,such2017deep,conti2018improving,wang2019paired,stanley2019designing,cully2015robots}. %\jc{add Go-Explore, add Stanley et al. Nature Machine Intelligence 2019 paper, consider adding 2015 Nature Robots that Adapt Like Animals}.
%Massive compute becomes one of key components that power some recent AI success. A recent study at OpenAI revealed that, \rw{add OpenAI ref} since 2012 the amount of compute used in the largest AI training runs has been increased exponentially with a 3.5 month-doubling time. 
In classic use cases such as supervised learning, distributed training is often straightforward to set up thanks to established machine learning frameworks (e.g.\ TensorFlow \cite{abadi2016tensorflow}, PyTorch \cite{paszke2017pytorch} and Horovod \cite{sergeev2018horovod} for deep learning applications, %\jc{should we show some love to Horovod here?}
and Spark MLlib \cite{meng2016mllib} outside deep learning). However,
%emergent approaches such as 
RL 
%\jc{we just did a cite dump for RL above and PB. I recommend creating the dump at the first instance of both. put all cites from here up there (deduplicating of course) and then do not do cites here for either RL or PB}
and population-based methods pose unique challenges for reliability, efficiency, and flexibility that frameworks designed for supervised learning fall short of satisfying.

%HERE

First, RL and population-based methods are typically applied in a setup that requires frequent %\jc{is continuous accurate? Frequent is.} 
interaction with simulators to evaluate policies and collect experiences, such as ALE \cite{bellemare2013arcade}, Gym \cite{brockman2016openai}, and Mujoco \cite{todorov2012mujoco}.
%\cite{bellemare2013arcade,brockman2016openai,todorov2012mujoco}
While neural network computation can leverage specialized hardware (e.g.\ GPUs and TPUs), the dominant computing workloads are often from simulations that only run on CPUs, and different simulation rollouts can take significantly different lengths of time to finish.  
%\ks{different from what?  do you mean that different rollouts take significantly different lengths of time to finish?}\rw{yes. modified}. 
This setup requires a distributed computing framework not only to leverage the large amount of computation similar to its counterpart in supervised learning, but also to handle its heterogeneity in resource usage. %, large variation in granularity and execution time among tasks, all due to the integration with simulators. %\jc{we have not explained why the integration with simulators leads to fault tolerance issues...or more generally why such issues arise at all in RL/PB}. 

Second, an ideal distributed computing framework should dynamically allocate resources for workloads whenever possible to ensure the maximum job throughput given a finite-size pool of computing resources. %\jc{after this sentence I thought you were referring to allocation within a single job for a single user}
A naive but natural choice could be to allocate computing resources according to the 
peak resource 
needed among all stages of computation.
%job priority specified by user.\jc{but now I am wondering. It's odd for a user to give priority within a job...so you must mean what? within a set of jobs for one user? or different users within a lab? I think this should be made explicitly clear up front}
However, for some RL and population-based methods, a better, more fine-grained dynamic scaling strategy is required to address variable computation needs at different phases of an algorithm.
For example, Go-Explore \cite{ecoffet2019go} requires only CPUs during its exploration phase, but relies on GPUs later in the robustification phase. Another example is POET \cite{wang2019paired}, whose execution could benefit from gradually scaling up resources according to the increasing size of active populations in the open-ended search. 

Third, such a distributed computing framework should keep a unified user interface consistent 
across a wide variety of backends, enabling practitioners to effortlessly turn a prototype algorithm that runs on a laptop into a high-performance distributed application that runs efficiently on a multi-core workstation, over a cluster of machines, or even on a public cloud, all with relatively few additional lines of code. This flexibility would maximize developmental efficiency and allow users to best utilize all the computing resources available to them.
%\ks{If we have to cut maybe the text below could be cut -- it seems to be more about Fiber specifically than what an ideal distributed system would do in general.}\rw{agree}\jc{if we don't have to cut, I think it is fine/good}
Moreover, the user interface should ideally be kept close to a familiar Python interface, which helps reduce adoption overhead. A good analogy in deep learning frameworks is PyTorch \cite{paszke2017pytorch}, which keeps many of its APIs close to those in NumPy \cite{oliphant2006guide}.

Among existing machine learning frameworks, those for supervised learning are not designed to address these new kinds of requirements as they do not naturally support simulation. Therefore, researchers and practitioners in RL traditionally resort to building one-off systems \cite{nair2015massively,silver2016mastering,tian2017elf,espeholt2018impala} for their specialized use cases, imposing a prohibitive systems engineering burden. More recently, RL frameworks have been developed directly based on specific deep learning frameworks, emphasizing support for either applications or research. For example, PyTorch-based Horizon \cite{gauci2018horizon} provides an end-to-end RL flow optimized for high performance on production data for real-life applications that do not require the level of flexibility for quick iteration needed in algorithm development in academic research.
%\jc{what evidence leads us to this claim? Would Edoardo and Jason disagree? How can the reader evaluate if this is true without providing the rationale that leads to the conclusion, instead of this (strongly worded, negative) conclusion?}\rw{I read their code. horizons are made for real-world application with highly optimized performance for processing production data and off-policly learning. I also read their paper. they clearly want to keep Horizons for industrial application than for from RL academic research,  not for algorithm development in research. but i agree that the wording here might be too strong...i made some changes}\jc{I like it! Nice work} 
Tensorflow-based Dopamine \cite{castro2018dopamine} instead retains enough flexibility for RL research, but lacks support for distributed training.

Ray \cite{moritz2018ray} tries to provide a generalized solution for emerging AI applications including RL.
%One exception \jc{that the above is true with only one exception is a pretty strong claim that could irk reviewers if not true...make sure you are right or rephrase...sounds like even with RLlib there is an argument to be made, so avoid such clear/strong/bold language}is Ray \cite{moritz2018ray},
%which targets emerging AI applications including RL.
%\jz{Talked to Thang, he thinks Ray is a more general framework, not just targeting at RL, maybe we should also mention RLlib which is for RL}.  
%Together with RLlib \cite{liang2017ray}\jc{unclear. do these software packages interoperate? ore are we now saying there are two software packages that are exceptions to the above rule}, 
It provides end-to-end distributed model training and serving.  %\jc{what does serving mean in this context? do we need to explain that?}\jz{serving means taking a trained model do inference on it, I think this should be clear to system for ML community} 
However, its design is influenced by Apache Spark \cite{zaharia2016apache} and adopts some memory-intensive approaches (e.g. task dependency graphs, control and object stores) to support an internal task scheduler. 
This approach can 
%thereby\jc{for me (an outsider to systems), nothing in the previous sentence makes the following claim obvious, so thereby doesn't seem appropriate. Consider cutting `thereby'}
be heavy-weight for algorithm development. Also, installing and deploying Ray on different backend platforms puts the burden of different customization on its users \cite{Ray_doc}.
%\jc{this sentence is not great English. Do you mean ``Also,  installing Ray on different backend platforms is difficult because of the customization required.}\rw{yes. agree. modified.}
%\jc{this claim...(see below)} \rw{refer to ray four different installation guide/setup for four different backends} %\url{https://ray.readthedocs.io/en/latest/}
%\jz{From Thang, maybe we need to give some example backends here, I'll think about it}
Beyond machine-learning-focused solutions, general-purpose parallel computing systems such as IPyParallel \cite{perez2007ipython} and OpenMPI \cite{boku2004openmpi} provide little direct support and no dynamic scaling for distributed training for RL and population-based methods. Ad-hoc solutions built on them are often not portable from one platform to another and are difficult to scale, significantly limiting the efficiency of algorithm development.%\jc{and these claims one has to take on faith. there is no cite, or specific fact or argument that I can evaluate the truth of} \rw{refer to the result section, the first result}\rw{add a few word that dynamic scalling is not supoorted}

%\rw{describe the challenges}
%\rw{Using a lot of compute is not easy!  especially we move beyond classic supervised learning.    }

To address these challenges %\jcsout{ head-on},
in this paper we introduce Fiber,
%\footnote{We will release the Fiber source code upon publication.}
a scalable distributed computing framework for RL that aims to achieve both flexibility and efficiency while supporting both research and practical applications: %and population-based methods that aims to address all the requirements described above, while retaining development efficiency and flexibility for both research and practical applications: 

(1) Fiber is built on a classic master-worker programming model, and introduces a \emph{task pool} as a lightweight but effective strategy to handle the scheduling of tasks.
%\jc{we have not explained what this means. do we need to?   } 
%\ks{an effective solution to handle the dependency?  or to handling the dependency?  it's not clear what we are saying is being solved

(2) Fiber leverages cluster management software (e.g. Kubernetes; \citeauthor{burns2016borg} \citeyear{burns2016borg}) %across different platforms for %\ks{what meaning does the word ``actual'' add here?}\rw{agree. deleted} 
for job scheduling/tracking. % and to ensure fault tolerance.

(3) Fiber does not require pre-allocating resources and can dynamically scale up and down on the fly, ensuring maximal flexibility. 
%\jc{pretty key. should we explain earlier or here that MPI cannot handle this issue, and how key this is for both fault-tolerance and dynamic scaling??}\rw{addressed earlier}

(4) Fiber is designed with maximizing development efficiency in mind 
%\jz{}
%implemented as an extension to Python's multiprocessing library 
so the user can effortlessly migrate from  multiprocessing on one machine to complete distributed training across multiple machines. 
The architecture of Fiber and experiments demonstrating its advantages are presented in this paper.

\section{Background}

This section briefly reviews RL and population-based methods, the targeted applications of Fiber, followed by a review of core concepts and components behind Fiber, e.g.\ multiprocessing, containers, and cluster management systems.

Confusingly, the words ``reinforcement learning'' (RL) can refer to a class of machine learning problems, the field of research about solving those problems, and a specific subset of algorithms that can solve those problems. 
The class of problems concerns how agents take actions in a (often simulated) environment to maximize a notion of cumulative reward. Without labelled input/output pairs, the focus of RL is to continually find a balance between exploration (of the space of possible actions) and exploitation (of the data currently gathered)  within an uncertain environment based on delayed and infrequent feedback.
RL algorithms are typically based on temporal-difference learning, and include the Q-learning and policy gradient families of algorithms \cite{sutton2018reinforcement}.
\emph{Population-based methods} are an additional class of search algorithms that can solve RL problems. They maintain a population of candidate solutions wherein encouraging behavioral diversity is a central drive.
They have produced state-of-the-art results in robotics \cite{cully2015robots,salimans2017evolution} and some hard-exploration RL problems \cite{ecoffet2019go}. Representative examples includes novelty search \cite{lehman2011abandoning} and Quality-Diversity algorithms \cite{lehman2011evolving,mouret2015illuminating,cully2015robots,pugh2016quality,wang2019paired,nguyen2016understanding,huizinga2018evolving}. %\jc{recommend adding CMOEA and Understanding Innovation Engines}, and open-ended search \cite{wang2019paired}\jc{add O'reiley article}. 
RL and population-based methods pose unique challenges to distributed training that Fiber aims to address.

\emph{Multiprocessing} is a Python standard library for parallel computing. It makes it easy for users to create new processes and create a pool of workers towards which tasks can be distributed. It is designed to leverage the computational power of modern multi-core CPUs on a single machine. Fiber follows the same interface as multiprocessing while extending many multiprocessing components to make them work in a distributed environment (i.e.\ across many computers in a computer cluster).

\emph{Cluster management systems} manage computer clusters and many machines simultaneously. They are the ``operating system''
%\ks{notice right before this comment the correct way to write quotes in latex -- do not use ".."}
layer on top of computer clusters and allow other applications to run on top of them. Examples include Apache Mesos \cite{hindman2011mesos}, Kubernetes \cite{burns2016borg}, Uber Peloton \cite{uberpeloton} and Slurm \cite{yoo2003slurm}. %\jc{add slurm?}

\emph{Containers} are a method of virtualization that package an application's code and dependencies into a single object. The aim is to allow application to run reliably and consistently from one environment to another environment. Containers are often used together with cluster management systems.

\begin{figure}
  \centering
  \includegraphics[width=1.0\linewidth]{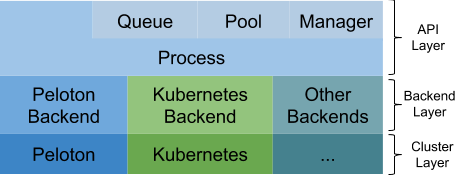}
  %\fbox{\rule[-.5cm]{0cm}{4cm} \rule[-.5cm]{4cm}{0cm}}
  %\end{center}
  \caption{\textbf{Fiber architecture.}}
  \label{fig:architecture}
\end{figure}

\section{Approach: Fiber}
Fiber provides users the ability to write applications for a large computer cluster with a %\jc{the or a?} 
standard and familiar library interface. 
%The section introduces Fiber and explains how it can can be used for RL algorithms and population-based methods. 
This section covers Fiber's design and applications.
%\jc{s}\jc{only adopt this suggestion if there are more than one error handling mechanims}.

\subsection{Architecture}

Fiber bridges the classical multiprocessing API with a flexible selection of backends that can run on different cluster management systems. To achieve this integration, Fiber is split into 3 different layers: the \textbf{API layer}, \textbf{backend layer} and \textbf{cluster layer}. %\jc{these are not in the figure. shouldn't they be?}.
The \textbf{API layer}
%provides basic building blocks for Fiber like Process, Queue, Pool and Manager types.
provides basic building blocks for Fiber like processes, queues, pools and managers.
They have the same semantics as in multiprocessing, but are extended to work in distributed environments. The \textbf{backend layer} handles tasks like creating or terminating jobs on different cluster managers. When a new backend is added, all the other Fiber components (queues, pools, etc.)
%\ks{new: above we capitalize terms like Queue and Pool, but here we use lower-case. we should be consistent and follow standard usage for these terms in this context within publications (which I don't know)} \jz{Rethink about this, terms like pools and queues should be lower case. When I first refer them, I call them Process, Queue, Pool and Manager \textbf{types}, which is just the Python class name of them. I can change the sentence to "... provides basic building blocks for Fiber like processes, queues, pools and managers." if this is less confusing }\rw{I like your proposal}\jz{fixed}
do not need to be changed. Finally, the \textbf{cluster layer} consists of different cluster managers. Although they are not a part of Fiber itself, they help Fiber to manage resources and keep track of different jobs, thereby reducing the number of items that Fiber needs to track. This overall architecture is summarized in figure \ref{fig:architecture}. 

\subsection{Fundamentals}

Fiber introduces a new concept called \emph{job-backed processes}. It is similar to the \emph{process} in Python's multiprocessing library, but more flexible: while a process in multiprocessing only runs on a local machine, a Fiber process can run remotely on a different machine or locally on the same machine. When starting a new Fiber process, Fiber creates a new job with the proper Fiber backend on the current computer cluster. Fiber uses containers to encapsulate the running environment of current processes, including all the required files, input data, and other dependent program packages, etc., to ensure everything is self-contained. All the child processes are started with the same container image as the parent process to guarantee a consistent running environment.
Because each process is a cluster job, its life cycle is the same as any job on the cluster.
To make it easy for users, Fiber is designed to directly interact with computer cluster managers. Because of this, Fiber doesn't need to be set up on multiple machines or bootstrapped by any other mechanisms, unlike Spark or IPyParallel.
It only needs to be installed on a single machine as a normal Python pip package. 

\subsection{Components}

Fiber implements most multiprocessing APIs on top of Fiber processes including pipes, queues, pools and managers.
%\ks{new: inconsistent capitalization again}\rw{addressed in an earlier reply} \jz{Now all these terms are in lower case now}

\begin{lstlisting}[language=Python, label=lst_code1, style=mystyle,caption={Fiber API Example}]
import random
import fiber

def worker(p):
    return random.random()**2 + random.random()**2 < 1

def main():
    NUM_SAMPLES = int(1e7)
    # fiber.Pool manages a list of distributed workers
    pool = fiber.Pool(processes=4)
    count = sum(pool.map(worker, range(0, NUM_SAMPLES)))
    print("Pi is roughly {}".format(4.0 * count / NUM_SAMPLES))

if __name__ == "__main__":
    main()

\end{lstlisting}

These components are critical to implement RL algorithms and population-based methods. An example code of the Fiber API are listed in code example \ref{lst_code1}. % \ks{All the code snippets are giving the term ``Listing \#'' -- is ``listing'' really a term people use formally for source code? If so it's okay, but maybe something more like ``Code Example \#'' would sound more familiar.} \jz{Agreed. I think ``listing'' is not very clear. I removed all the ``listing'' and just reference code examples with section numbers}

\textbf{Supported multiprocessing components}. %\ks{If we need more space these headings could be removed and the text could just speak for itself, perhaps with a little modification for continuity.}
\textit{Queues and pipes} in Fiber behave the same as in multiprocessing. The difference is that queues and pipes are now shared by multiple processes running on different machines.
Two processes can read from and write to the same pipe. Furthermore, queues can be shared between many processes on different machines and each process can send to or receive from the same queue at the same time. %\ks{Previously you capitalized Queue and Pipe, but here it is lower case.  Find out the correct usage and then change all mentions to use the correct one consistently. (My instinct is that lower case is preferred here because these are common terms/words, at least for this community.)} \jz{Agreed and fixed}
Fiber's queue is implemented with Nanomsg\footnote{https://nanomsg.org/}, %\cite{hintjens2013zeromq}, 
a high-performance asynchronous message queue system. 
\textit{Pools} are also supported by Fiber. They allow the user to manage a pool of worker processes. Fiber extend pools with \textit{job-backed processes} so that it can manage thousands of (remote) workers per pool. %\ks{capitalization is again inconsistent, this time for ``pool.''}.
Users can also create multiple pools at the same time.
\textit{Managers and proxy objects} enable multiprocessing to support shared storage, which is critical to distributed systems. Usually, this function is handled by external storage like Cassandra \cite{lakshman2010cassandra}, Redis \cite{carlson2013redis}, etc. Fiber instead provides built-in in-memory storage for applications to use. The interface is the same as multiprocessing's Manager type. %\ks{should Manager be capitalized? should multiprocessing? need to be careful throughout this paper on capital vs. lower-case}\jz{fixed}.
In this way, Fiber provides a shared storage that is convenient to use and high performance. %\ks{The last sentence of this paragraph is not grammatical, appears to be a run-on sentence, and also incoherent (I don't know what it's trying to say).  Needs to be rewritten carefully.} \jz{fixed}

\textbf{Unsupported multiprocessing components.}
\textit{Shared memory} is used heavily by frameworks like PyTorch \cite{paszke2017pytorch} and Ray \cite{moritz2018ray}. In general, it can improve performance of inter-process communications on the same machine. However, it is not available when communicating over computer network, which is common for distributed systems. Thus Fiber provides managers and proxy objects as the primary means to share data instead.  %\ks{The intended point here is unclear. The title above says ``not \emph{yet} supported,'' which implies that we think it should or will be supported in the future, but the language in this paragraph implies that it doesn't need to be supported, so ``not yet'' seems like the wrong message.  What are we really trying to say here?  If it's not needed, then why not just say that directly so the reader isn't confused?} \jz{fixed}
\textit{Locks} can be very import for coordinating between different processes and preventing race conditions. However, in a distributed environment, it may cause wasting large amount of computation resources.
%locks are often not as useful as when running locally because the storage system usually handles race conditions on its own.
Therefore, we excluded locks from the supported APIs as it's not needed by most RL and population-based methods. %\ks{Same issue as the paragraph above -- if we don't need them then why are we saying it's not ``yet'' supported?  Will it ever be supported?  Why would we ever support it if it's not needed?}\jz{fixed}

\subsection{Applications}
%\jz{this section has been completely rewritten}

RL and population-based methods are two major applications for Fiber.
These algorithms generally require frequent interactions between policies (usually represented by a neural network), %which decide what action to take given a specific state
and environments (usually represented by a simulator like ALE \cite{bellemare2013arcade}, OpenAI Gym \cite{brockman2016openai}, and Mujoco \cite{todorov2012mujoco}). %which decide what are the instantaneous rewards and state transitions under current actions.
%People usually choose to put policies and environments on different machines to improve efficiency. Thus, making sure the communication is efficient between these components is critical to the efficiency of the algorithm.
The communication pattern for distributed RL and population-based methods usually involves sending different types of data between machines: actions, neural network parameters, gradients, per-step/episode observations and rewards, etc. Actions can be either discrete (represented by an integer) or continuous (represented by a float number). The number of actions that needs to be transmitted are usually less than a thousand. The size of observations and neural network parameters can be larger than actions. Depending on the neural network used, the size of parameters (and gradients) can range from bytes \cite{mnih2015human} to megabytes \cite{OpenAI_dota}.
%\jc{this should be openAI et al. and have a year}\jz{fixed}
%Although gigabyte-sized neural networks do exist \cite{radford2019language}, they are not common in RL and population-based learning.
%\jc{yet! but likely will be. either add this, or remove the whole sentence: is it necessary?}\jz{There are some recent work which gets transformers working with RL. So I'm more inclined to remove this sentence.}

Fiber implements pipes and pools to transmits these data. Under the hood, pools are normal Unix sockets, providing near line-speed communication for the applications using Fiber. Modern computer networking usually has bandwidth as high as hundreds of gigabits per second. Transmitting smaller amount of data over a network is usually fast \cite{dean2007software}. Fiber can run each simulator in a single process and communicate actions, observations, rewards, parameters or gradients via pipes between different processes. If the size of parameters or gradients is too large, Fiber can be used together with Horovod \cite{sergeev2018horovod}, which leverages GPU to GPU communication for faster communication.
Additionally, the inter-process communication latency does not increase much if there are many different processes sending data to one process because data transfer can happen in parallel. This fact makes Fiber's pools suitable for creating the foundation of many RL and population-based learning algorithms because simulators can run in each pool worker process and the results can be transmitted back in parallel.

%\ks{italics makes sense when introducing a new term, but once already introduced it's not clear why we continue to italicize}\jz{fixed}
\vspace{-0.11in}
\begin{lstlisting}[language=Python, label=lst_code2,caption={ES implemented with Fiber}]
def evaluate(theta):
    # do rollout and return the reward
    return reward

def train():
    N = 1000
    workers = 10
    # fiber.Pool manages a list of distributed workers
    pool = fiber.Pool(workers)
    theta = init()
    for i in range(n):
        noises = [sample_noise() for i in range(N)]
        thetas = [theta + sigma * noise for noise in noises]
        rewards = pool.map(evaluate, thetas)
        s = sum([rewards[i] * noises[i] for i in range(N)])
        theta = alpha/(workers * sigma) * s
\end{lstlisting}

The distinction between pool- and pipe-based communication is that pools usually ignore task order while pipes keep order. For pools, each task (neural network evaluation, etc.) can be mapped to any of the worker processes. This is suitable for algorithms like ES \cite{salimans2017evolution} or POET \cite{wang2019paired}, where each task is stateless. Each evaluation can run on any of the pool worker processes and only the (per episode) end results need to be collected back in parallel. An example of ES algorithm implemented with Fiber is listed in code example \ref{lst_code2}. On the other hand, pipes can maintain the order of each task. Each simulator is mapped to a fixed process so that worker processes can maintain their internal state after each step. At each step, each worker process only needs to accept actions that are for that specific worker and send the results (instantaneous rewards, state transitions, etc.) back. This makes it suitable for RL algorithms like A3C \cite{mnih2016asynchronous}, PPO \cite{schulman2017proximal}, etc. An example of RL implemented in Fiber is listed in code example \ref{lst_code3}.

\begin{lstlisting}[language=Python, label=lst_code3,caption={RL implemented with Fiber}]
# fiber.BaseManager is a manager that runs remotely
class RemoteEnvManager(fiber.BaseManager):
    pass

class Env(gym.env):
    # gym env
    pass

RemoteEnvManager.register('Env', Env)

def build_model():
    # create a new policy model
    return model

def update_model(model, observations):
    # update model with observed data
    return new_model

def train():
    model = build_model()
    manager = RemoteEnvManager()
    num_envs = 10

    envs = [manager.Env() for i in range(num_envs)]

    obs = [envs[i].reset() for i in num_envs]
    for i in range(1000):
        actions = model(obs)
        obs = [env.step() for action in actions]
        model = update_model(model, obs)
\end{lstlisting}

\subsection{Scalability}

There are two key considerations on scalability: (1) how many resources a framework can manage and (2) how many resources an algorithm can use. Because Fiber relies on the cluster scheduler to manage the resources including CPU cores, memory and GPUs, there is little role for Fiber in managing the resources except in tracking started processes and properly terminating them when computation is completed. Fiber schedules each task at most once. When batching is enabled, multiple tasks can be scheduled at the same time to improve efficiency.
%Fiber can scale up to the whole computer cluster by removing unnecessary computation from the critical path. \ks{previous sentence is very unclear and relies on undefined concepts.  what is ``unnecessary computation,'' what is the ``critical path,'' and how is the unnecessary computation removed?}
It is also possible to run Fiber across multiple clusters, but the network communication cost could make Fiber less efficient. 
% scale up and down
% Resource preemption
Because Fiber does not require pre-allocating resources, it can scale up and down with the algorithm it runs. Compared to static allocation, Fiber can return unused resources back to the cluster when they are not needed. Furthermore, when it needs more resources, it can ask the cluster manager for more resources. This approach makes it suitable for algorithms that runs heterogeneous tasks in different stages.

\subsection{Error Handling}
%Error handling
% pool failure

%\begin{wrapfigure}[17]{o}{0.4\textwidth}
%  \centering
%  \includegraphics[width=1.0\linewidth]{pool_failure.png}
  %\fbox{\rule[-.5cm]{0cm}{4cm} \rule[-.5cm]{4cm}{0cm}}
  %\end{center}
%  \caption{Pool failure case} %\ks{some fonts in this figure (``Pending table'') are too small for the reader to see at printed size}}
%  \label{fig:pool_failure}
%\end{wrapfigure}

\begin{figure}
  \centering
  \includegraphics[width=1.0\linewidth]{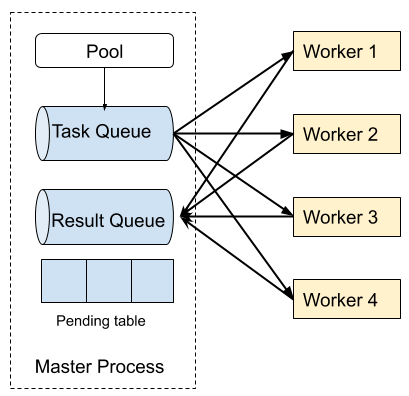}
  %\fbox{\rule[-.5cm]{0cm}{4cm} \rule[-.5cm]{4cm}{0cm}}
 \caption{\textbf{Task execution failure handling.} Fiber implements task queues, result queues and pending tables. 
 %\ks{missing a caption here.  reader needs some explanation of the figure.  note that readers sometimes skim figures before reading text, or without reading text at all.  nevertheless, should be succinct and to the point.}\jz{Added a short description here} 
  When a task is fetched from the task queue, an entry is added to the pending table. Tasks successfully executed are added to the result queue, while failed tasks are resubmitted to the task queue according to the corresponding entry in the pending table.}
  \label{fig:pool_failure}
\end{figure}

Fiber implements pool-based error handling (Figure \ref{fig:pool_failure}). When a new pool is created, an associated task queue, result queue and pending table are also created. Newly created tasks are then added to the task queue, which is shared between the master process and worker processes. Each of the workers fetches a single task from the task queue, and then runs task functions within that task. Each time a task is removed from the task queue, an entry in the pending table is added. Once the worker finishes that task, it puts its results in the result queue. The entry associated with that task is then removed from the pending table.

If a pool worker process fails in the middle of processing, that failure is detected by the parent pool that serves as the process manager of all the worker processes.  Then the parent pool puts the pending task from the pending table back into task queue if the previously failed process has a pending task. Next, it starts a new worker process to replace the previously failed process and binds the newly created worker process to the task queue and the result queue.
%\ks{is there anything interesting or special about all this?  it's not clear what is or is not unique about how Fiber handles errors.  what about these details is interesting?}

\section{Experiments}

Fiber is designed to scale the computation of algorithms like RL and population-based methods easily. In this section, we evaluate Fiber on three different tasks to show its benefits: \textit{Framework overhead} is tested on a dummy workload, \textit{Evolution Strategies (ES)} experiments show its potential for population-based training, and \textit{Proximal Policy Optimization (PPO)} experiments test the same for RL. Results show that Fiber has low overhead and can easily scale ES to thousands of CPU workers. This is a big improvement compared to IPyParallel which can only scale to hundreds of CPU workers.
%\jc{in a vacuum this number is meaningless. is that a lot or a little? how does that compare to what other packages can do, and/or what has been possible historically? You need to provide context to these sorts of numbers.}\jz{Added a sentence "This is a big improvement compared to IPyParallel which can only scale to hundreds of CPU workers.".}.
Also, Fiber can easily reuse existing code like OpenAI Baselines \cite{dhariwal2017openai} and seamlessly expand PPO to use hundreds of distributed environment workers. OpenAI baselines does not support computation in such scale.
%\jc{same comment: is that a lot? a little?}\jz{Added "OpenAI baselines does not support computation in such scale"}
In Addition, it only requires a few lines of changed code.

\subsection{Framework overhead} \label{framework_overhead}

\begin{figure*}
     \centering
     \begin{subfigure}[b]{0.3\textwidth}
         \centering
         \includegraphics[width=1.0\textwidth]{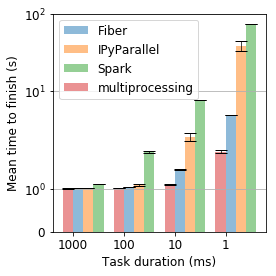}
         %\fbox{\rule[-.5cm]{0cm}{3cm} \rule[-.5cm]{3cm}{0cm}}
         \caption{Framework overhead}
         \label{fig:framework_overhead}
     \end{subfigure}
     \hfill
     \begin{subfigure}[b]{0.3\textwidth}
         \centering
         \includegraphics[width=1.0\linewidth]{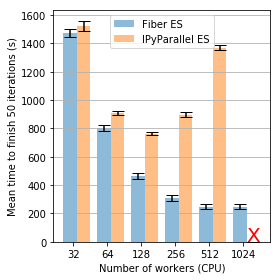}
         \caption{Evolution Strategies}
         \label{fig:es_ipp_fiber}
     \end{subfigure}
     \hfill
     \begin{subfigure}[b]{0.3\textwidth}
         \centering
         \includegraphics[width=1.0\linewidth]{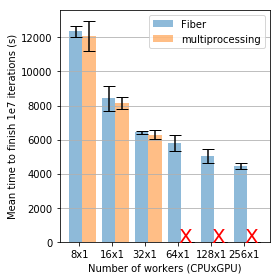}
         \caption{PPO}
         \label{fig:ppo_ipp_fiber}
     \end{subfigure}
        \caption{\textbf{Results.} (a) %\ks{These should be sentences. This caption has very poor writing style -- everything is written exactly the same way}\jz{fixed}
        Different frameworks are compared on mean time to finish a batch of tasks with different task durations. The optimal finishing time is 1 second. Note that the y-axis is in log scale. (b) Fiber scales better than IPyParallel when running ES with different numbers of workers. Each worker runs on a single CPU. (c) While Fiber performs on par with multiprocessing with 8, 16 and 32 workers (CPUs), it scales beyond a single machine (32 CPUs) and still provides performance improvements.
        %When running PPO algorithm with different number of workers, Fiber's performance matches with multiprocessing which only works locally. Fiber scales beyond a single machine and still provides performance improvements. 
        All workers are connected to a master process with a single GPU. An red X in the figure means the framework did not finish the test due to internal errors.}
        %\jc{explain why there are red x's for both b and c. do this both here and in the main text (at the places I mention in the main text)}\jz{added}
        %\ks{The figure needs a caption so you could read it and make some sense out of it even if you skipped the main section text.  A sentence or two here should explain what all these graphs are meant to convey to the reader.  Also, I think the fonts in the graphics are likely too small to see when at printed size.}}
        \label{fig:three graphs}
\end{figure*}

%\jz{Add Spark setup, the relation between Spark and PySpark}
The aim of this test is to probe how much overhead the framework adds to the workload. For this purpose, we compare Fiber, Python multiprocessing library, Spark, and IPyParallel. The testing procedure is to create a batch of workload that takes a fixed amount of time in total to finish. The duration of each single task ranges from 1 second to 1 millisecond. We run five workers for each framework locally and adjust the batch size to make sure the total finish time for each framework is roughly 1 second (i.e.\ for 1 millisecond duration, we run 5,000 tasks). Results are in figure \ref{fig:framework_overhead}.

We use multiprocessing as a reference because it is very lightweight and does not implement any additional features beyond creating new processes and running tasks in parallel. Additionally, it exploits communication mechanisms only available locally (e.g.\ shared memory, Unix domain socket, etc.), making it difficult to be surpassed by other frameworks that support distributed runs resource management across multiple machines and cannot exploit similar mechanisms. It thus serves as a good reference on the performance that can be expected.
%\jc{It thus serves as an upper bound on the performance that can be expected. The question is how close each framework can get to this upper bound of performance.}\jz{TODO: I want to say it's a good reference point but not an upper bound because we have some new results which shows we can surpass multiprocessing. What would be a better term for this? benchmark?}
Fiber shows almost no difference when task durations are 100ms or greater, and is much closer to the multiplrocessing than the other frameworks as the task duration drops to 10 or 1ms.
%The results show that Fiber performs similarly to multiprocessing for each task duration
%\jc{that's just not true when I look at the plots. If we say things that seem wrong, we lose the trust of the reader. You may have an argument that the difference is so small as to not matter, but that is different than saying there is no difference. For the 1ms Task Difference, there is definitely a difference. Can we instead say something like Fiber shows almost no difference when task durations are 100ms or greater, and is much closer to the multiplrocessing than the other frameworks as the task duration drops to 10 or 1ms...or some other more factual claim? }\jz{fixed}
The small difference in performance is a reasonable cost to gain the ability to run on multiple machines and scale to the whole computer cluster.
%Although Fiber removes significant overhead for communication, unlike multiprocessing, it cannot exploit communication mechanisms only available locally (e.g.\ shared memory, unix domain socket, etc.). Nevertheless, even with this limitation, Fiber's performance is not far from multiprocessing. 
Compared to Fiber, IPyParallel and Spark fall well behind at each task duration. When the task duration is 1 millisecond, %\ks{inconsistent abbreviation -- above you call it 1 millisecond but here 1}\jz{fixed}
IPyParallel takes almost 8 times longer than Fiber, and Spark takes 14 times longer. This result highlights that both IPyParallel and Spark introduce considerable overhead when the task duration is short, and are not as suitable as Fiber for RL and population-based methods, where a simulator is used and the response time is a couple of milliseconds.

%\ks{Something is missing in this subsection: all it basically tells us is that Fiber is a bit worse than multiprocessing.  But why is that okay?  We need to remind the reader why there are other advantages that make up for or justify why Fiber is important if it performs worse (even just a little worse) than the alternative.  It shouldn't take a lot of text, but it needs to be said explicitly -- do not expect the reader to figure it out for themselves.  Something like (doesn't have to be verbatim this), ``The small difference in performance compared to multiprocessing is a reasonable cost to gain the ??? that Fiber provides.''}

\subsection{Evolution Strategies (ES)}

To probe the scalability and efficiency of Fiber, we compare it here exclusively with IPyParallel because Spark is slower than IPyParallel as shown above, and multiprocessing does not scale beyond one machine.  %\ks{When you say Spark is slower than IPyParallel do you mean because the results showed Spark is significantly slower than IPyParallel?  Or are you implying it is a well known fact that Spark is slower?} \jz{I mean the result above showed that Spark is slower than IPyParallel. I'll change the narrative}.
We evaluate both frameworks on the time it takes to run 50 iterations of ES (figure \ref{fig:es_ipp_fiber})
to test the scalability and efficiency of both frameworks. %\jc{non-sequitur}\jz{fixed}
%because\jc{non-sequitur}ES is a highly parallelizable algorithm \cite{salimans2017evolution}.\jz{fixed}
With the same workload, we expect Fiber to finish faster because it has much less overhead than IPyParallel as shown in the previous test. %that with the same workload, Fiber should finish quicker because it has much less overhead than IPyParallel as shown in the previous test.
%We implemented two versions of the ES algorithm \cite{salimans2017evolution},
%one in Fiber and the other in IPyParallel. 
%\ks{why is it logical that here we are comparing to IPyParallel but in the previous section we compare to multiprocessand Spark? The reader won't make these leaps on their own -- needs to be spelled out explicitly.  For everything you do in experiments, we need to tell the reader WHY.}
For both Fiber and IPyParallel, the population size of 2,048, so that the total computation is fixed regardless of the number of workers. The same shared noise table trick mentioned in \citeauthor{salimans2017evolution} (\citeyear{salimans2017evolution}) is also implemented in both. Every 8 workers share one noise table.  %The testing environment is a customized version of BipedalWalker environment. 
%\rw{should we report the difference between this env and the standard BipedalWalker env? like this: The experimental domain in this work is a modified version \cite{POET_bipedal} of the ``Bipedal Walker Hardcore'' environment of the OpenAI Gym \cite{brockman2016openai}. }\ks{Yes, the current text fails to explain what BipedalWalker is or where it comes from.} 
The experimental domain in this work is a modified version of the ``Bipedal Walker Hardcore'' environment of the OpenAI Gym \cite{brockman2016openai} with modifications described in \citeauthor{POET_bipedal} (\citeyear{POET_bipedal}).

%\ks{Okay but why?  What are we trying to figure out here? What is the hypothesis?  Should Fiber be better for some reason?  For example, we could say something like ``The question is, ...''}
The main result is that Fiber scales much better than IPyParallel and finishes each test significantly faster. The length of time it takes for Fiber to run gradually decreases with the increase of the number of workers from 32 to 1,024. In contrast, the time for IPyParallel to finish \emph{increases} from 256 to 512 workers. IPyParallel does not finish the run at 1,024 workers due to communication errors between its processes (hence the red X in figure \ref{fig:es_ipp_fiber}).
%\jc{ (hence the red X in Fig.~\ref{INSERT})}.\jz{fixed}
This unexpected failure undermines the ability for IPyParallel to run large-scale parallel computation.
%\ks{why not? is that expected?  sounds mysterious}.
Overall, Fiber's performance exceeds IPyParallel for all numbers of workers tested. Additionally, unlike IPyParallel, Fiber also finishes the run with 1,024 workers. This result highlights Fiber's better scalability compared to IPyParallel.
% IPyParallel to finish for all number of workers tested. \ks{is that expected?  does it confirm a hypothesis? does it follow from some theory of what is better about Fiber?  results are reported here in a vacuum.}

%\begin{figure}
%  \centering
%  %\fbox{\rule[-.5cm]{0cm}{3cm} \rule[-.5cm]{4cm}{0cm}}
%  \includegraphics[width=0.4\linewidth]{fiber_ipp_es.png}
%  \caption{Evolution Strategies}
%  \label{fig:es_ipp_fiber}
%\end{figure}

% ES + box2d
% ES + mujoco
% baselines + PPO
\subsection{Proximal Policy Optimization}

%\begin{figure}
%  \centering
%  %\fbox{\rule[-.5cm]{0cm}{3cm} \rule[-.5cm]{4cm}{0cm}}
%  \includegraphics[width=0.4\linewidth]{fiber_ipp_ppo.png}
%  \caption{PPO}
%  \label{fig:ppo_ipp_fiber}
%\end{figure}

%Unlike other platforms, we don't need to implement our own version of Proximal Policy Optimization (PPO) \cite{schulman2017proximal} for this test. 

To assess Fiber's suitability for RL, we want to see how difficult it is to run a typical RL algorithm in a distributed setup. 
%It is well known that converting a single-machine multiprocessing implementation of a RL algorithm to a code that can utilize multiple machines for scalable distributed training typically requires significant engineering effort.
It is well known that parallelizing a single-machine multiprocessing implementation of RL algorithm requires significant engineering effort \cite{heess2017emergence}.
%Converting an RL algorithm that work on a single-machine to an algorithm that utilizes multiple machines for scalable distributed training typically requires significant engineering effort \cite{heess2017emergence}.
However, Fiber makes it as simple as changing \textit{one line} of code. No other platform as far as we know offers this capability.  
%assess how suitable Fiber is for RL, we want to see how difficult it is to run a typical RL algorithm in a distributed setup. In addition, we want to determine whether adding more resources helps with the training time. 
To demonstrate this simplicity, we chose a widely-used multiprocessing implementation of the popular PPO algorithm \cite{schulman2017proximal} from OpenAI baselines \cite{dhariwal2017openai}, and converted it to code that can run over hundreds of machines by simply replacing \texttt{import multiprocessing as mp} with \texttt{import fiber as mp}. 
%, and not using replay buffer which make it more complex to parallelize because multiple workers can add experiences to the replay buffer with linear speed-up. There have been efforts to distribute PPO , but it requires modifying and reimplementing the algorithm. . 
%To adapt code for a single-machine PPO to run distributed PPO, one only needs to start with the  change \emph{one line} of code, and run PPO on a computer cluster in a distributed fashion. More specifically, the modification only requires  In this way, 

We then compare the performance of the distributed version of PPO enabled by Fiber with its original multiprocessing implementation on Breakout in the Atari benchmark \cite{bellemare2013arcade} with a total of 10 million frames for training. The test runs on one 1080 Ti GPU for the neural network policy and a variable number of CPU workers running OpenAI Gym \cite{brockman2016openai} environments. We run 8 to 32 (maximum CPU cores available on our test machine) workers for multiprocessing and 8 to 256 workers for Fiber. As shown in figure \ref{fig:ppo_ipp_fiber}, Fiber scales beyond 32 workers. When running 64 and more workers, its performance beats the best result multiprocessing can get from a single machine. With 256 workers, the total time by Fiber is less than half of that with 8 workers. These results show that Fiber can scale RL beyond local machines. Additionally, when running a small number of workers, Fiber virtually matches the performance of multiprocessing because Fiber has low overhead. There is only 1\% to 3\% difference between Fiber and multiprocessing. This observation is significant because multiprocessing leverages optimizations only available locally as noted previously.
%in section \ref{framework_overhead}. 
%\ks{new:note that the section reference above does not work because sections in this style have no numbers! should find a different way to refer to the section, or just say "previously" if it's too awkward.}\jz{Good catch!, I changed it to "previously"}
%Unlink multiprocessing, Fiber can take PPO to a much scale not limited by the resource available on a single machine. 
Finally,
%Amdahl's law \cite{amdahl1967validity} provides the theoretical speedup in latency of task execution when the workload is fixed.
the PPO implementation in OpenAI baselines has 2 major time consuming parts: the environment step and the model step.
%We run the model step on a single GPU for both  and run the environment step either on CPUs on a single machine (multiprocessing) or multi machines (Fiber).
We noticed sub-linear speedup on both multiprocessing and Fiber due to the fact that only the environment step can be benefited from adding more workers. This is a limitation in the current OpenAI baselines implementation.
\section{Conclusion}

In this work, we presented a new distributed framework that allows efficient development and scalable training. Experiments highlight that Fiber achieves many goals, including efficiently leveraging a large amount of heterogeneous computing hardware,
%\jc{how is this point different from the next point? is this heterogeneous over time (in which case it is the same point as the next point) or over hardware (GPUs and CPUs)? try to make this clearer}, \jz{changed to "computing hardware"}
dynamically scaling algorithms to improve resource usage efficiency, reducing the engineering
%\jc{engineering}\jz{added} 
burden required to make 
%\jc{required to make}\jz{added}\jcsout{of programming} 
RL and population-based 
%\jc{RL and population-based}\jz{added} 
algorithms work on computer clusters, %distributedly, \ks{``learning programming distributedly'' is ambiguous and poorly worded (``distributedly'' is not an English word).  do you mean the burden on the programmer of learning to program for a distributed system?  or do you mean allowing learning algorithms to work on a distributed system?}
and quickly adapting to different computing environments to improve research efficiency. At the same time, Fiber outperforms existing frameworks like IPyParallel and Spark. Finally, while Fiber is designed for RL and population-based learning, its general API in principle allows it to be applied in much broader contexts. %With Fiber, anything that uses multiprocessing can be easily adapted to run on a distributed computer cluster by changing just a few lines of code. 
%To conclude, Fiber provides a new and easy platform to efficiently develop, train and scale RL and population-based methods, as well as other distributed applications. 
We expect it will further enable progress in solving hard RL problems with RL algorithms and population-based methods by making it easier to develop these methods and train them at the scales necessary to truly see them shine \cite{clune2019ai}.
%\jc{not necessary, but you could put an AI-GA reference here so people know what we mean by this without adding any more text}\jz{added}

\bibliographystyle{aaai}
\bibliography{ref}

\end{document}